\documentclass[letterpaper, 10 pt, conference]{ieeeconf}  

\IEEEoverridecommandlockouts                              

\usepackage{times} 
\usepackage{amsmath} 
\usepackage{amssymb}  
\usepackage{bm}

\usepackage{amssymb}
\usepackage{graphicx}
\usepackage{stackengine}
\usepackage{float}
\usepackage{subcaption}
\usepackage{hyperref}
\title{\LARGE \bf
Fast Online Optimization for Terrain-Blind Bipedal Robot Walking with a Decoupled Actuated SLIP Model}

\author{Ke Wang$^{1}$, Hengyi Fei$^{2}$ and Petar Kormushev$^{1}$
\thanks{$^{1}$Ke Wang and Petar Kormushev are with Robot Intelligence Lab,Dyson School of Design Engineering, Imperial College London, UK {\tt\small k.wang17@imperial.ac.uk}}%
\thanks{$^{2}$Hengyi Fei is with the Department of Electrical \& Electronic Engineering, Imperial College London, UK}%
}
\begin{document}

\maketitle
\thispagestyle{empty}
\pagestyle{empty}

\begin{abstract}
We present a highly reactive controller which enables bipedal robots to blindly walk over various kinds of uneven terrains while resisting pushes. The high level motion planner does fast online optimization for footstep locations and Center of Mass (CoM) height using the decoupled actuated Spring Loaded Inverted Pendulum (aSLIP) model. The decoupled aSLIP model simplifies the original aSLIP with Linear Inverted Pendulum (LIP) dynamics in horizontal states and spring dynamics in the vertical state. The motion planning can be formulated as a discrete-time Model Predictive Control (MPC) and solved at a frequency of 1k~HZ. The output of the motion planner using a reduced-order model is fed into an inverse-dynamics based whole body controller for execution on the robot. A key result of this controller is that the foot of the robot is compliant, which further extends the robot's ability to be robust to unobserved terrain changes. We evaluate our method in simulation with the bipedal robot SLIDER. Results show the robot can blindly walk over various uneven terrains including slopes, wave fields and stairs. It can also resist pushes while walking on uneven terrain.

\end{abstract}

\section{INTRODUCTION}

To make bipedal robots really suitable for many applications, it is important that they can go out of the lab and walk in the complex real world environment. Real world environments contain various kinds of uneven terrains: slopes, wave fields, stairs, etc. Most existing controllers that allow a bipedal robot to walk over uneven terrains require predefined footstep locations or exact information about the terrain height changes \cite{mordatch2010robust}, \cite{englsberger2015three}, \cite{liu2015trajectory}. However, even with most advanced sensors there are some uncertainties on the perception of the terrain. In contrast, humans can easily walk on uneven terrains, such as outdoor environments. without extra thought or careful planning. Therefore to have a reactive controller that is robust to unobserved uneven terrain changes is important.

The Spring Loaded Inverted Pendulum (SLIP) model has become a popular model for walking and running in the legged robotic research \cite{raibert1986legged}. Despite its simplicity, it has been proven to capture essential dynamics properties of walking and running \cite{geyer2006compliant}. The standard setting of SLIP model is energy-conservative: it assumes there is no energy loss at impact. Though this assumption simplifies the control analysis, it doesn't resemble the reality. There is energy loss on physical systems, and robots \cite{ahmadi2006controlled} \cite{cassieRobot} designed to approximate SLIP dynamics have added actuation to compensate for the energy dissipation. As a result, the actuated Spring Loaded Inverted Pendulum (aSLIP) model \cite{ernst2010spring} is proposed for a better approximation of the real robot dynamics. The aSLIP model has been successfully used to design controllers not only for SLIP-like robots \cite{apgar2018fast} \cite{green2020planning} \cite{xiong2018bipedal} but also as a template model for humanoid robots on uneven terrain walking \cite{liu2016terrain}.

An important step in making a controller reactive is to achieve real-time constrained optimization. However, due to the nonlinear dynamics that arise from the 3D aSLIP model, fast optimization is difficult. \cite{liu2016terrain} used a gait synthesized from a library of gaits acquired from off-line optimization, but this requires a large computation load offline and cannot cover all possible situations. \cite{apgar2018fast} decoupled 3D aSLIP model to facilitate fast computation, but the continuous dynamics of decoupled aSLIP model is still nonlinear and there is no theoretical guarantee of fast convergence. \cite{xin2019react} used the simpler Linear Inverted Pendulum (LIP) model to design a reactive controller for flat ground walking, but this model cannot be applied to walking on uneven terrains.
\begin{figure}[t]
\centering
\includegraphics[width=0.8\columnwidth]{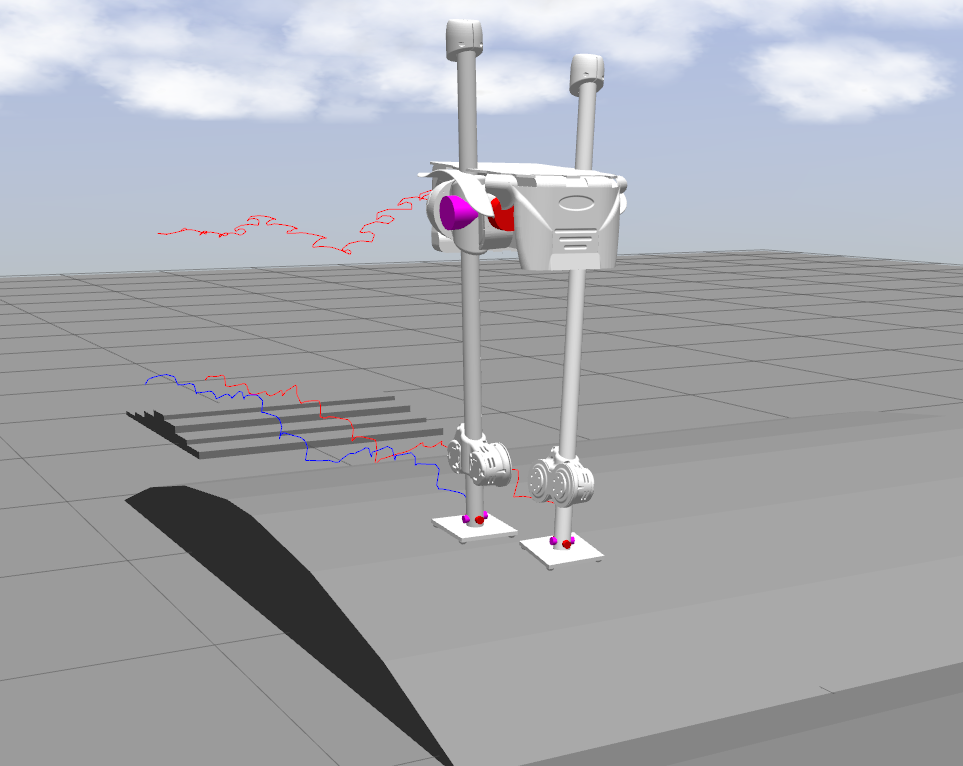}
\caption{SLIDER robot walks on uneven terrain. The three traces are trajectories of CoM, left foot and right foot respectively.}
\label{fig:slider_uneven}
\vspace{-5mm}
\end{figure}

Our paper proposes a reactive controller that enables robots to blindly walk over uneven terrains by optimizing horizontal footstep locations and center of mass (CoM) height online. Under the assumption that the angle the modelled inverted pendulum makes with the vertical is relatively small, we can decouple the 3D aSLIP model into a 1D actuated spring model responsible for $z$ direction and 2D LIP model responsible for $x$ and $y$ directions. The dynamics of all three dimensions can be written as linear equations in a discrete-time state space manner. We formulate the online step planner as a discrete-time model predictive control (MPC) problem and solve it by Quardratic Programming (QP). To facilitate fast computation, the spring length is constrained to change linearly. As a result, we get a step planner which runs at a frequency of 1000~HZ. The step planner using a simple model is embedded into the inverse-dynamics based whole body controller \cite{Herzog_2015} \cite{kim2019highly} which tackles the inconsistency between the simple model used in high level planning, and full robot dynamics. With the whole body controller, the feet of the robot show great compliance and this helps the robot to transition between different terrains without any information about the terrain. Due to the fast execution frequency and compliance of the foot, our proposed controller enables the robot to blindly walk over various kinds of moderately uneven terrains including slopes, wave fields and stairs. Our controller can also handle disturbances from all direction while walking on uneven terrain. We validated our controller on the straight-legged bipedal robot SLIDER \cite{wang2020slider} in Gazebo simulation. \\

The main contribution of the paper is the reactive controller which enables bipedal robots to blindly walk over various uneven terrains while resisting disturbances. With reasonable assumptions, we can decouple the nonlinear 3D aSLIP dynamics and optimize the footstep location and CoM height separately. By proper reformulation, the control of the vertical dynamics can be formulated as QP and therefore a solution is guarenteed to be found. Our controller is simple to implement and computationally efficient; the optimization of both vertical and horizontal motions can be solved by QP and executed at 1k~HZ\footnote{Video summary: \url{https://youtu.be/ROyV-ZP8dxA}}.

\section{SYSTEM OVERVIEW}
\subsection{The SLIDER robot}

\begin{figure}[t]
\centering
\includegraphics[width=0.8\columnwidth]{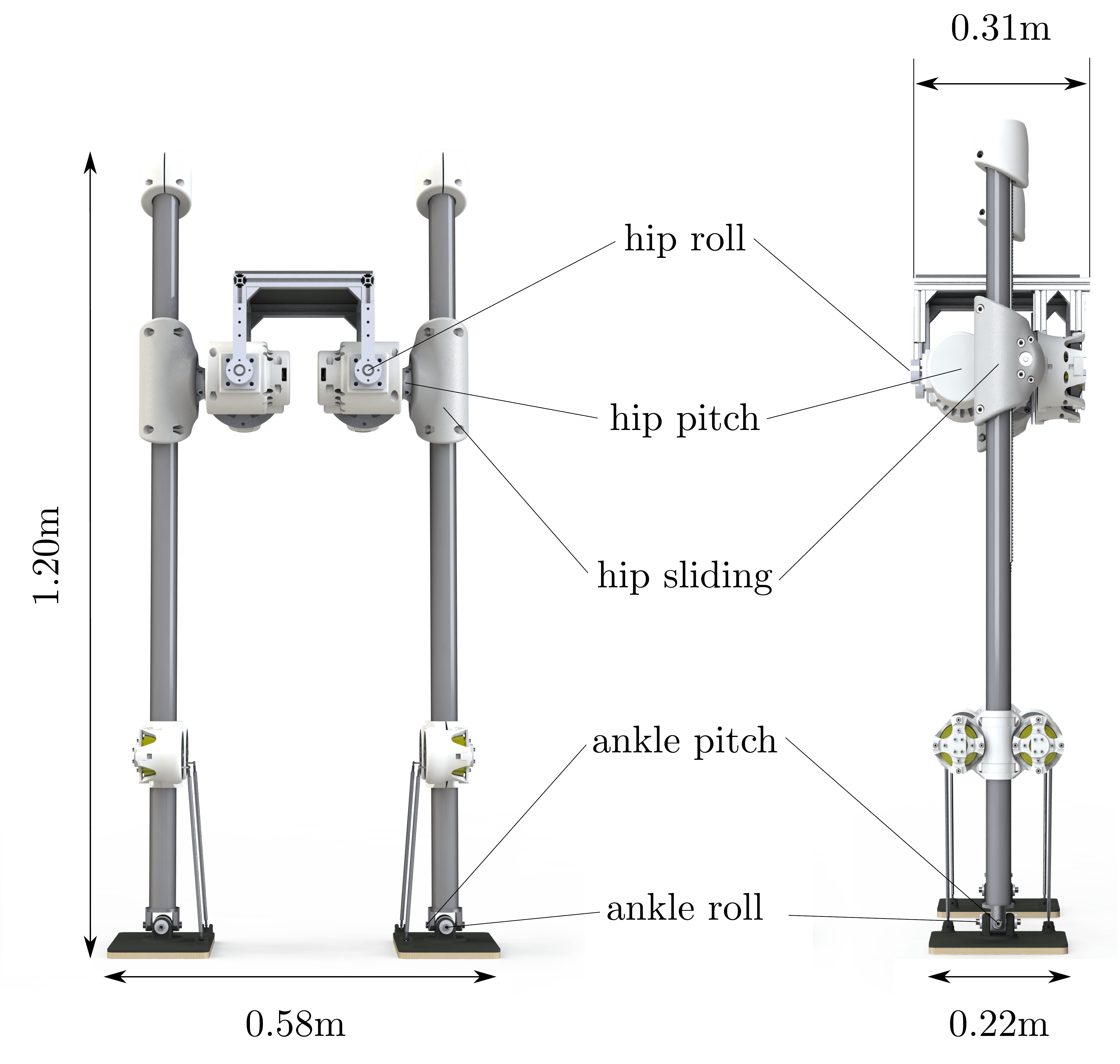}
\caption{The dimension and joint configuration of the SLIDER robot.}
\label{fig:slider_dim_config}
\vspace{-5mm}
\end{figure}

 {SLIDER} is a knee-less bipedal robot designed by the Robot Intelligence Lab at Imperial College London, as shown in Fig. \ref{fig:slider_dim_config}. SLIDER is 1.2~m tall and has 10 Degrees of Freedom (DoF), namely hip pitch, hip roll, hip slide, ankle roll and ankle pitch on each leg. The robot is very lightweight (14.5~kg in total) and most of its weight is concentrated at the pelvis. The legs are made of carbon fiber reinforced polymer and each leg weights only 0.4~kg. The prismatic knee joint design is an unique feature of this robot that differentiate it from many other robots with anthropomorphic design. Due to its sliding mechanism and lightweight leg design, SLIDER can be well approximated with an aSLIP model which can greatly simplify the planning and control problem. Moreover the lightweight leg with a large range of motion makes the robot suitable for agile locomotion.  
 
\subsection{The Control Hierarchy}
\begin{figure} 
\centering 
\includegraphics[width=1.0\columnwidth]{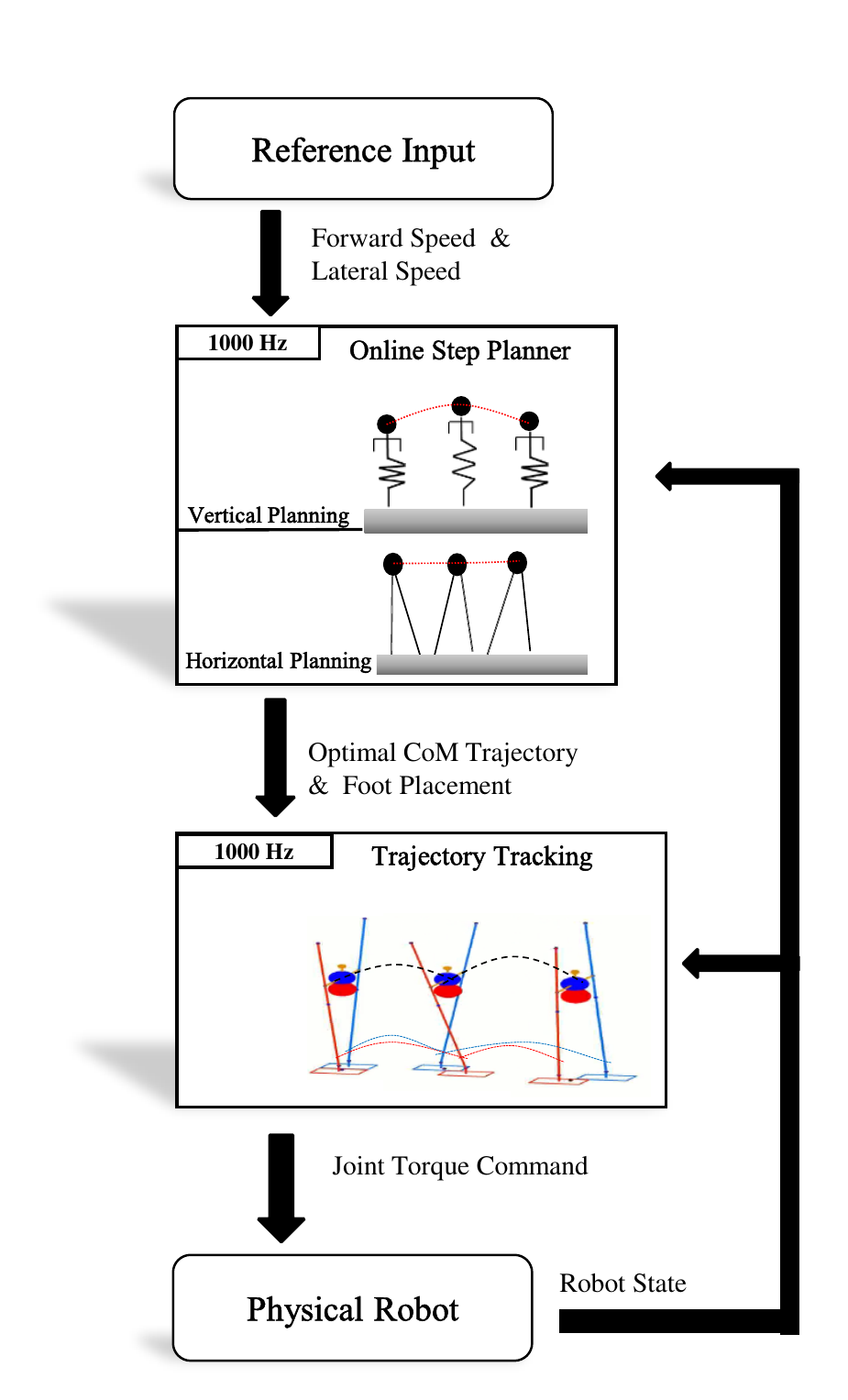}
\caption{The hierarchical controller with the execution frequency of each level. The high level trajectory planner takes the desired velocity as input and generates the optimal CoM trajectory along with the foot placement. The low level whole body controller \cite{Herzog_2015} \cite{feng2015optimization} considers full dynamics of the robot and tracks the trajectory at a frequency of 1k~HZ.} 
\label{fig.main2} 
\vspace{-5mm}
\end{figure}

The controller presented has a hierarchical structure as shown in Fig \ref{fig.main2}. Due to the computational complexity of using full body motion planning, a high level planner optimizes online only the foot placement in $x$, $y$ directions and the CoM height using the decoupled aSLIP model. The low level whole body controller \cite{Herzog_2015} \cite{feng2015optimization} tracks the trajectory generated by high level planner. The whole body controller considers the full dynamics of the robot and generates the consistent torque command for each joint.

\section{Online Footstep Planning}

Motion planning with the full dynamics of the robot is too computationally expensive to be executed at a fast frequency. Instead we use reduced order models for online footstep planning: the LIP model is used to plan footsteps in the horizontal plane and the 1D actuated spring model is used generate  in the vertical direction. 
\subsection{The Decoupled aSLIP Model}
\begin{figure} [b] 
\centering 
\includegraphics[scale=0.6]{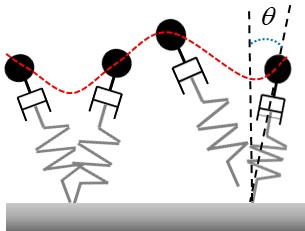}
\caption{The aSLIP model. This model combines the SLIP model with a virtual linear actuator. $\theta$ is the angle the inverted pendulum makes with the vertical.} 
\label{fig:aSLIP} 
\end{figure}

The aSLIP model is different from the classical SLIP model in that the aSLIP model is a combination of the SLIP model and a virtual linear actuator, as shown in Figure \ref{fig:aSLIP}. Introducing such a virtual actuator gives the aSLIP model the capability to actively modify the reference spring length during each walking phase and makes aSLIP model better at handling vertical changes than the SLIP model.
However, similar to the SLIP model, the dynamics of aSLIP model presented above is nonlinear because of the coupling in vertical and horizontal dynamics, and the exact dynamic equation needs to be numerically integrated. An approximate model which decouples the CoM states into horizontal and vertical directions will be used in this paper. This decoupled approximate model simplifies the nonlinear dynamics of the original aSLIP model and makes the fast online optimization of footstep locations and CoM height possible. To make the approximation valid, two assumptions are made. The first assumption is that the vertical deviation of CoM over each walking period is small relative to the height of the CoM, such that individual steps can be modelled using the LIP model. The second assumption is that the angle the inverted pendulum makes with the vertical is small, such that the spring contributes to the CoM's vertical behavior only. This is reasonable, considering a typical small step length is small compared to the height of the CoM. These assumptions are summarised in Fig. \ref{fig:aSLIP}.
\begin{figure}[t]
\begin{subfigure}{0.6\textwidth}
    \includegraphics[width=0.7\linewidth]{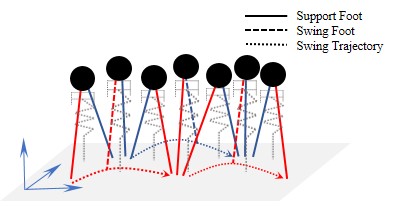}
\end{subfigure}
\vspace{0.3cm}

\begin{subfigure}{0.22\textwidth}
  \includegraphics[width=0.7\linewidth]{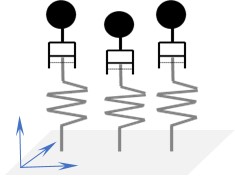}
\end{subfigure}
\hspace{-0.5cm}
\begin{subfigure}{0.22\textwidth}
  \includegraphics[width=0.7\linewidth]{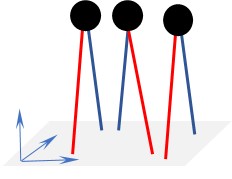}
\end{subfigure}
\label{fig:decoupled}
\caption{The decoupled aSLIP model during walking (top), with vertical dynamics (bottom left) and horizontal dynamics (bottom right).}
\vspace{-5mm}
\end{figure}

\subsection{Vertical Dynamics of decoupled aSLIP Model}
The dynamics of the decoupled aSLIP model in vertical direction can be calculated as \ref{eq1}, where $z$ denotes position of CoM in vertical direction, $r$ denotes the reference spring length, and $k$ is the stiffness of the spring.
\begin{equation}
    m \ddot{z} = - mg + k (z-r)  \tag{1}
\label{eq1}
\end{equation}
 Compared with the classical SLIP model, the reference spring length $r(t)$ is treated as a time-varying optimization variable, which is contributed by the virtual linear actuator. We also constrain $r(t)$ to change linearly between initial and final values over the phase to facilitate fast computation. Therefore, the reference spring length should satisfy the condition below,
\begin{equation}
    r(t) = r_0 +\frac{t}{T} (r_T-r_0) \tag{2}
    \label{eq2}
\end{equation}
where, $r_0$ and $r_T$ represents the reference spring length at the start and at the end of one step respectively, $T$ is the step duration. We can formulate the dynamic equation (\ref{eq1}) into a state space equation as,
\begin{equation}
\dot{Z} = \left[\begin{array}{cc}
     0& 1 \\
     -\omega_z^2&0 
\end{array}
\right] Z +\left[\begin{array}{cc}
     0 \\
     \omega_z^2 
\end{array}
\right](r-\frac{g}{\omega_z^2}) \tag{3a}
    \label{eq3}
\end{equation}
where the states of CoM in vertical direction are defined as $Z = [z, \dot{z}]^T$ and we define $\omega_z =\sqrt{\frac{k}{m}} $. Further, we can denote $r-\frac{g}{\omega_z^2}$ as $u_z$, and equation (\ref{eq4}) can be written into a linear state space equation as,
\begin{equation}
\dot{Z} = \left[\begin{array}{cc}
     0& 1 \\
     -\omega_z^2&0 
\end{array}
\right] Z +\left[\begin{array}{cc}
     0 \\
     \omega_z^2 
\end{array}
\right]u_z \tag{3b}
    \label{eq4}
\end{equation}
We discretize equation (\ref{eq4}) with sampling time $T_s$ and can obtain,
\begin{equation}
  Z_{k+1} = \Phi(T_s)Z_k+\int_0^{T_s} \Phi(T_s) dt \left[\begin{array}{cc}
     0 \\
     \omega_z^2 
\end{array}
\right]u_{z,k}  \tag{4a}
\end{equation} 
where,
\begin{equation}
\Phi(T_s) =  \left[\begin{array}{cc}
     cos(\omega_z T_s)& sin(\omega_z T_s)/\omega\\
    \omega_z \cdot sin(\omega_z T_s)&cos(\omega_z T_s)
\end{array}
\right] \tag{4b}
\end{equation}
\begin{equation*}
    \int_0^{T_s} \Phi(T_s)dt\left[\begin{array}{cc}
     0 \\
     \omega_z^2 
\end{array}
\right] = \left[\begin{array}{cc}
     1-cos(\omega_z T_s) \\
     \omega_z \cdot sin(\omega_zT_s)
\end{array}
\right] \tag{4c}
\label{eq4c}
\end{equation*}
 
We can also write this equation as:
\begin{equation*}
    Z_{k+1} = A_z(T_s)Z_k+B_z(T_s)u_{z,k} \tag{4d}
\end{equation*}

where $Z_k$ and $Z_{k+1}$ are CoM states in the vertical direction at time k and k+1 respectively, $A_z(T_s)$ and $B_z(T_s)$ are time-varying matrices, $u_{z,k}$ is the control input for the spring at time k. Therefore, $N$ future steps can be computed given fixed sampling time $T_s$:

\begin{equation}
\begin{array}{cc}
    Z_{1} = A_z(T_s)Z_0+B_z(T_s)u_{z,0} \\
    Z_{2} = A_z(T_s)Z_1+B_z(T_s)u_{z,1} \\
    ... \\
    Z_{N} = A_z(T_s)Z_{N-1}+B_z(T_s)u_{z,N-1} \tag{5a}
    \label{eq5a}
\end{array}
\end{equation}

where $N$ is the number of steps to be optimized which should be more than 1. In this formulation, the control input vector for the spring $\bar{U} = [u_{z,0}, u_{z,1}, ... u_{z,N-1}]$ is the optimization variable. Equation (\ref{eq5a}) can also be written in a matrix form for MPC:

\begin{equation}\bar{Z} =
    \left[\arraycolsep=1.2pt
    \begin{array}{c}
         A_z\\
         A_z^2\\
         ...\\
         A_z^N 
    \end{array}\right] Z_0 + \left[\arraycolsep=1.2pt
    \begin{array}{cccc}
        B_z & 0 &0 &0  \\
        A_zB_z &B_z &0 &0\\
        ... & ... & ... &...\\
        A_z^{N-1}B_z & A_z^{N-2}B_z &... &B_z
    \end{array}\right] \bar{U}_{z} \tag{5b}
    \label{eq5b}
\end{equation}
where $\bar{Z}$ and $\bar{U}_{z}$ are the vectors of states and control inputs in z direction. 

\subsection{Horizontal Dynamics of Decoupled aSLIP Model}

Under the assumption that the leg angle $\theta$ is relatively small and therefore the horizontal and the vertical dynamics can be decoupled, the horizontal dynamics then becomes a classical Linear Inverted Pendulum (LIP) model. Similar to \cite{xin2019react}, given the continuous dynamics of LIP model in the x direction,

\begin{equation}
    \ddot{x} = \frac{g}{z_0}(x-p_x)
    \tag{6}
\end{equation}

the discrete state space equation can be reformulated into a linear state space as below, with state defined as $X = [x,\dot{x}]^T$, and input $u_x$ defined as the footstep positions in x direction.
\begin{equation}
    X_{k+1}=A_x(T_s)X_k+B_x(T_s)u_{x,k} \tag{7a}
\end{equation}
Where
\begin{equation*}
    A_x(T_s) = \left[\begin{array}{cc}
     cosh(\omega_x T_s)& sinh(\omega_x T_s)/\omega_x\\
    \omega_x \cdot sinh(\omega_x T_s)&cosh(\omega_x T_s)
\end{array}
\right] \tag{7b}
\end{equation*}
\begin{equation*}
    B_x(T_s) = 
\left[\begin{array}{cc}
     1-cosh(\omega_x T_s) \\
     \omega_x \cdot sinh(\omega_xT_s)
\end{array}
\right] \tag{7c}
\end{equation*}
Similarly, states in the predicted time horizon can also be obtained by,
\begin{equation}\bar{X} = \left[\arraycolsep=1.2pt
    \begin{array}{c}
         A_x \\
         A_x^2 \\
         ... \\
         A_x^N 
    \end{array}
    \right] X_0 + \left[\arraycolsep=1.2pt
    \begin{array}{cccc}
        B_x & 0 &0 &0  \\
        A_xB_x &B_x &0 &0\\
        ... & ... & ... &...\\
        A_x^{N-1}B_x & A_x^{N-2}B_x &... &B_x
    \end{array}\right] \bar{U}_{x} \tag{8}
\end{equation}
where $\bar{X}$ and $\bar{U}_{x}$ are the vectors of states and control inputs in x direction. The dynamics in the y direction has identical formulation to the dynamics in the x direction.

\subsection{Foot Placement and CoM Trajectory Optimization}
The footstep planning can be formulated as a QP problem due to the fact that both horizontal and vertical dynamics are linear and the constraints are linear. The formulation is:

\begin{align}
  \min\limits_{u_{x}, u_{y}, u_{z}}  \Gamma \tag*{(cost function) \quad (9a)}  \\ 
   \textrm{s.t.} \quad X_{k+1}=A(T_s)X_k+B(T_s)u_{k} \tag*{(dynamics) \quad (9b)} \\
 h(\bold{p}_{j}) < 0 \tag*{(reachability)\quad (9c)} 
\end{align}

where $\Gamma =\Gamma_1+\Gamma_2+\Gamma_3$ is the cost function term, $X$ and $u$ are general representations of states and control inputs respectively and $\bold{p}_{j}$ is the left or right foot position.

The cost formulation of the vertical and horizontal control problems are both composed of the three parts: $\Gamma_1$, $\Gamma_2$ and $\Gamma_3$. The first part is minimizing the difference between the predicted state and the referenced state, which drives the CoM state to reach the desired one from the given current state. The formulation is:
\begin{equation}
    \Gamma_1 = \vert\vert X_N-X_N^{ref}\vert\vert^2_P+\sum_{k=0}^{N-1}\vert\vert X_k-X_k^{ref}\vert\vert^2_Q \tag{10}
\end{equation}
where $P$ and $Q$ are the weight matrix respectively. For the horizontal trajectory planning, we only care about the velocity of CoM in both directions, because we want a reactive footstep planner which is not constrained by a absolute reference trajectory. From another point of view, tracking the reference velocity is equivalent to tracking the relative CoM reference position. $X^{ref}$ is easy to define with the desired sagittal and frontal velocity.
However, the reference velocity in vertical direction is not constant during one step. To get the reference velocity, we can integrate equation (\ref{eq1}) and get the continuous dynamics equation:

\begin{equation}
    z(t) = d_1\cos{\omega_z t} + d_2\sin{\omega_z t} + r(t) - g/{\omega_z}^2 \tag{11a}
\end{equation}
where
\begin{align}
    d_1 = z_0 - r_0 + g/{\omega_z}^2 \tag{11b}\\
    d_2 = \dot{z}_0/{\omega_z}^2 - (r(t)-r_0)/(T\omega) \tag{11c}
\end{align}
In this equation, $z_0$ and $\dot{z}_0$ are the CoM vertical position and velocity at the start of the current step respectively. Then the time-varying reference velocity in z direction can be derived from the continuous dynamics.

The second part is minimizing the difference between the reference control input $u_k^{ref}$ and the optimized control input $u_{k}$ and R is the weight matrix. 
\begin{equation}
    \Gamma_2 = \sum_{k=0}^{N-1}\vert\vert u_k-u_k^{ref}\vert\vert^2_R \tag{12}
\end{equation}
For the reference control input in z direction, recall that the spring length is constrained to change linearly, as indicated by equation (\ref{eq2}). If we denote $\Delta r_{i}$ as the change of spring length during one sampling time $T_s$ at step $i$, then it is clear that  $\Delta r_{i}$ is constant throughout step $i$, and the reference control input in z direction  $\bar{U}^{ref}_{z}$ is therefore obtained as below,
\begin{equation}
       \bar{U}^{ref}_z = \left[\arraycolsep=1.2pt
    \begin{array}{c}
         r_0 I_{N_r} \\
         0_{N_s} \\
         ... \\
         0_{N_s} 
    \end{array}
    \right] +  \left[\arraycolsep=1.2pt
    \begin{array}{cccc}
        1 & 0 &...&0   \\
        1 & 1 &...&0 \\
        1 & 1 & ...&0 \\
        1 & 1&... &1
    \end{array}\right] \left[\arraycolsep=1.2pt
    \begin{array}{c}
         0_{N_r} \\
         \Delta r_{1} I_{N_s} \\
         ... \\
         \Delta r_{N_{steps}} I_{N_s} 
    \end{array}
    \right] \tag{13}
\end{equation}
where $N_r$ is the remaining number of sampling points for the current step and $N_s$ is the number of sampling points for one step. $N_{steps}$ is the number of predicted steps.

For the horizontal trajectory planning, the input of the system is the footstep position. The reference control input in x direction $\bar{U}^{ref}_x$ is then obtained according to the current support foot position $P_0$ and the difference between two consecutive steps $\Delta P_{x,i}$. 

\begin{multline}
\centering
       \bar{U}^{ref}_x = \left[\arraycolsep=1.2pt
    \begin{array}{c}
         I_{N_r} \\
         0_{N_s} \\
         ... \\
         0_{N_s} 
    \end{array}
    \right]P_{x,0} +\left[\arraycolsep=1.2pt
    \begin{array}{ccc}
         0_{N_r}&... & 0_{N_r} \\
         I_{N_s}&... & 0_{N_s}\\
         ... & ... &... \\
         0_{N_s}&... & I_{N_s}
    \end{array}
    \right]\left[\arraycolsep=1.2pt
    \begin{array}{cc}
        P_{x,1}   \\
        P_{x,2}   \\
        ...   \\
        P_{x, N_{steps}}   \\
    \end{array}
    \right]\tag{14}
\end{multline}
\begin{equation*}
    \left[\arraycolsep=1.2pt
    \begin{array}{cc}
        P_{x,1}   \\
        P_{x,2}   \\
        ...   \\
        P_{x,N_{step}}   \\
    \end{array}
    \right] = \left[\arraycolsep=1.2pt
    \begin{array}{c}
         I_{N_r} \\
         I_{N_s} \\
         ... \\
         I_{N_s} 
    \end{array}
    \right]P_{x,0}+\left[\arraycolsep=1.2pt
    \begin{array}{cccc}
        1 & 0 &0 &0  \\
        1 & 1 &0 &0\\
        ... & ... & ... &...\\
        1 & 1 &... &1
    \end{array}\right]\left[\arraycolsep=1.2pt
    \begin{array}{cc}
        \Delta P_{x,1}   \\
        \Delta P_{x,2}   \\
        ...   \\
        \Delta P_{x,N_{step}}   \\
    \end{array}
    \right]
\end{equation*}

The third part is tracking the desired change of the control input between two consecutive steps, with $d_i$ denoting the desired difference and $W$ denoting weights. For the vertical trajectory planning we set the desired difference to equal zero, this means the footstep planner tries to keep a constant CoM height during one step to make the decoupled aSLIP assumption valid. 
\begin{equation}
    \Gamma_3^z = \sum_{i=1}^{N_{steps}}\vert\vert\Delta r_i-d_i^z\vert\vert_W^2= \sum_{i=1}^{N_{steps}}\vert\vert\Delta r_i\vert\vert_W^2 \tag{15a}
\end{equation}
For horizontal trajectory planning,
\begin{equation}
    \Gamma_3^{x,y} = \sum_{i=1}^{N_{steps}}\vert\vert\Delta P_i^{x,y}-d_i^{x,y}\vert\vert_W^2 \tag{15b}
\end{equation}

In x direction, $d_i^x$ is the step length which is calculated by the desired speed multiplied by the step time. In y direction, $d_i^{y}$ is defined as $d_i^{y}= d_{step}*2*(-1)^j$, where $d_{step}$ is the desired inter-feet clearance distance, and $j$ is the flag for the supporting foot, 0 stands for left support and 1 stands for right support. This relative distance regularization term
is introduced to keep the feet away from each other to avoid self-collision. Morever, as this part of the cost function only includes relative distances, it helps to produce a reactive footstep planner which keeps the robot walking even when unexpected disturbance is applied.


The reachability constraint is responsible for making sure the footstep location is physically possible,

\begin{equation}
      \bold{L} - \bold{r}^{x,y} < \bold{p}_{j} - \bold{c} < \bold{L} + \bold{r}^{x,y} \tag{16}
\end{equation}

where $\bold{L}$ is the nominal offset of the foot position from the CoM of the robot, $\bold{r}^{x,y}$ is the reachability constraint in the x and y directions and $\bold{c}$ is the CoM position.

\section{Trajectory Tracking}
The low level trajectory tracking generates corresponding torques for each joint to minimize the difference between the actual body trajectory and the desired trajectory given by the high level trajectory planner. Since the trajectory planning does not include the full body dynamics, the dynamic inconsistency of the torque command generated by the high level planner is significant. The whole body controller solves the inverse dynamics based on the full robot dynamics.

\subsection{Rigid Body Dynamics}
The walking robot is modeled as a floating based rigid body system with coordinates $q=[q_{b},q_{r}]$. Here, $q_b \in \mathbb{R}^7$ represents the position and orientation of the floating base using quaternions, and $q_r \in \mathbb{R}^{10}$ represents the joint configuration. Inspired by \cite{Herzog_2015}, the full dynamics can be decomposed into an underactuated part and an actuated part:
\begin{equation}
    \begin{bmatrix}
    \bm{M}_f \\ \bm{M}_a 
    \end{bmatrix} \bm{\ddot{q}} +
    \begin{bmatrix}
        \bm{H}_f \\ \bm{H}_a
    \end{bmatrix} = 
    \begin{bmatrix}
        \bm{0} \\ \bm{S}_a
    \end{bmatrix} \bm\tau +
    \begin{bmatrix}
        \bm{J}^{\mathrm{T}}_f \\ \bm{J}^{\mathrm{T}}_a
    \end{bmatrix} \bm{f},
    \tag{16}
\end{equation}

where $\bm{M}$, $\bm{H}$, $\bm{S}_a$, $\bm {\tau}$, $\bm{J}$ and $\bm{f}$ are the mass matrix, the Coriolis force vector and the gravitation force vector, the actuator selection matrix, the joint torques vector, the stacked contact Jacobian and the reaction force vector respectively. The subscript $f$ and $a$, indicates the floating part and the actuated part respectively.

\subsection{Contact Force Constraint}
Friction is a important contributing factor to the stability of walking to prevent the foot from slipping. The criterion is that the contact force should satisfy the inequality below, 
\begin{equation}
    f_z \geq 0, \quad \sqrt{f_x^2+f_y^2} \leq \mu f_z \tag{17}
    \label{eq:friction}
\end{equation}
where $\mu$ is the friction coefficient. However, this formulation is a quadratic inequality constraint and rises the problem of constraining fast optimization. Therefore, we use the pyramidal friction model which is a linear inequality constraint for fast optimization:
\begin{equation}
    f_z \geq 0, \quad \vert f_x \vert \leq \frac{\mu f_z}{\sqrt{2}}, \quad \vert f_y \vert \leq \frac{\mu f_z}{\sqrt{2}} \tag{18}
\end{equation}
\begin{figure*}[t!]
\centering 
\begin{subfigure}{0.22\textwidth}
  \includegraphics[width=1.0\linewidth, height=2.5cm]{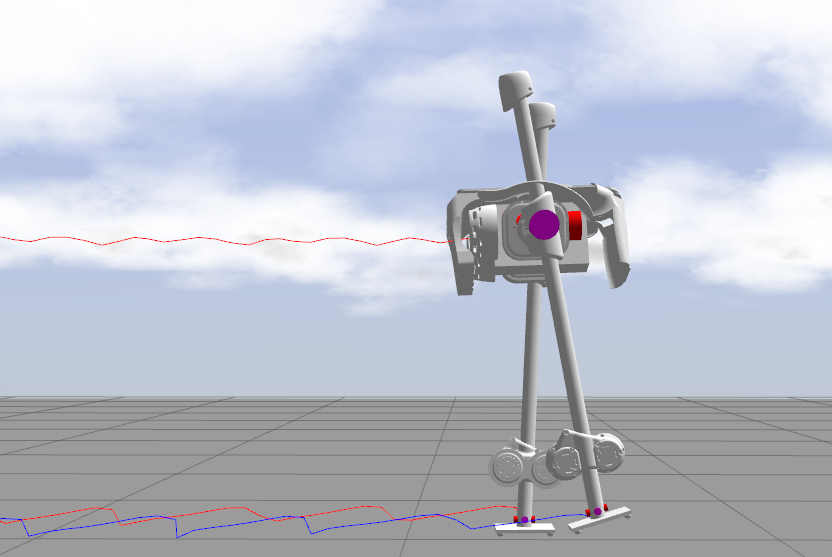}
  \caption{}
  \label{fig:uneven_a}
\end{subfigure}
  \hspace{0.01cm}
\begin{subfigure}{0.22\textwidth}
  \includegraphics[width=1.0\linewidth, height=2.5cm]{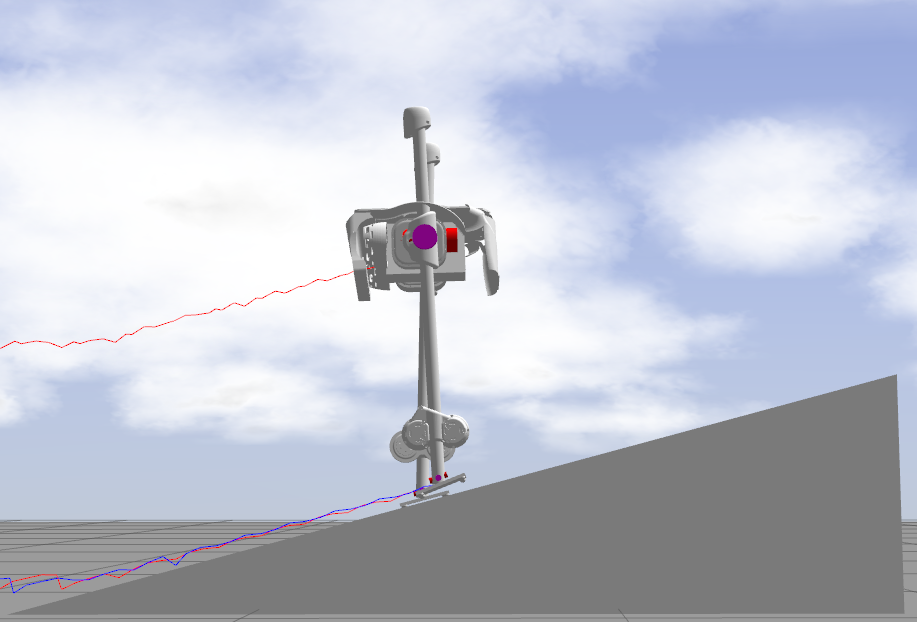}
  \caption{}
  \label{fig:uneven_b}
\end{subfigure} 
\begin{subfigure}{0.22\textwidth}
  \includegraphics[width=1.0\linewidth, height=2.5cm]{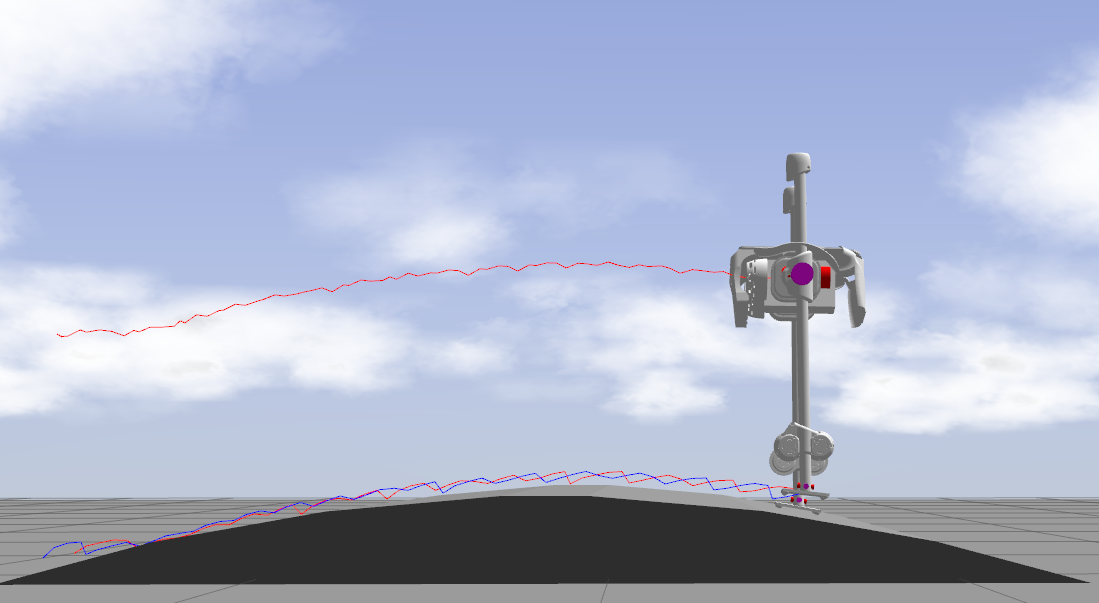}
  \caption{}
  \label{fig:uneven_c}
\end{subfigure}
  \hspace{0.01cm}
\begin{subfigure}{0.22\textwidth}
  \includegraphics[width=1.0\linewidth, height=2.5cm]{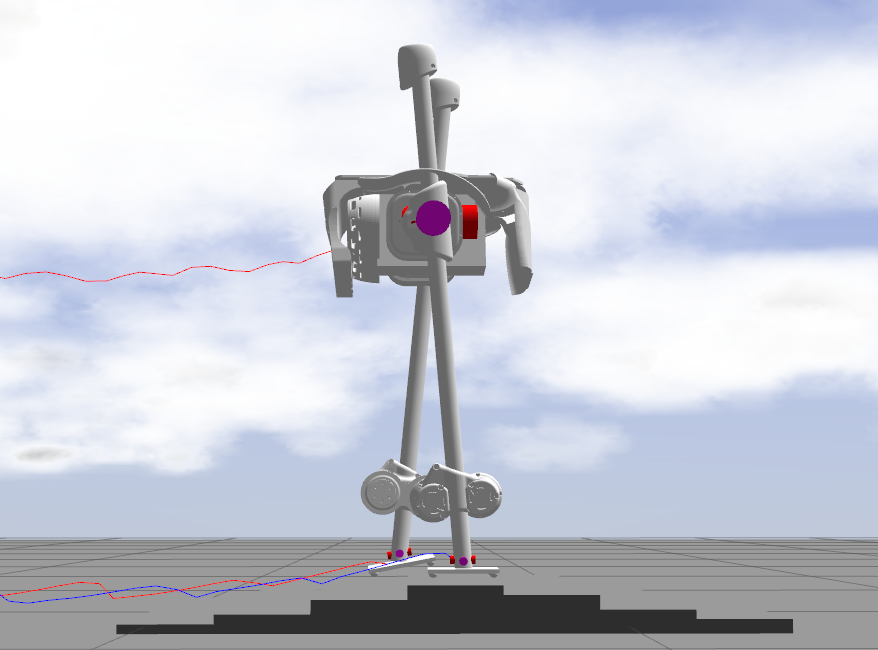}
  \caption{}
  \label{fig:uneven_d}
\end{subfigure} 

\begin{subfigure}{0.22\textwidth}
  \includegraphics[width=1.0\linewidth]{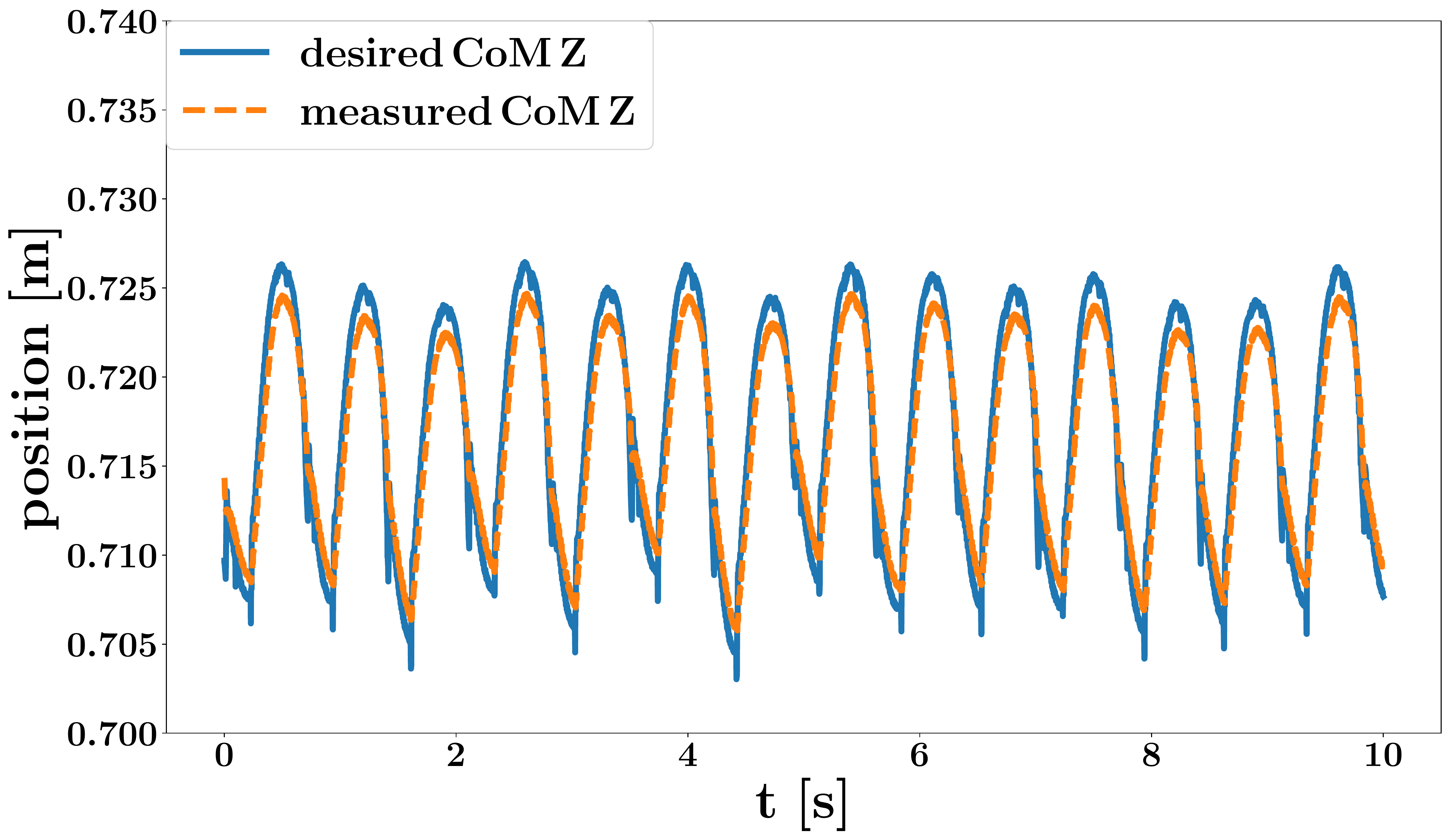}
  \caption{}
  \label{fig:uneven_e}
\end{subfigure}
  \hspace{0.01cm}
\begin{subfigure}{0.22\textwidth}
  \includegraphics[width=1.0\linewidth]{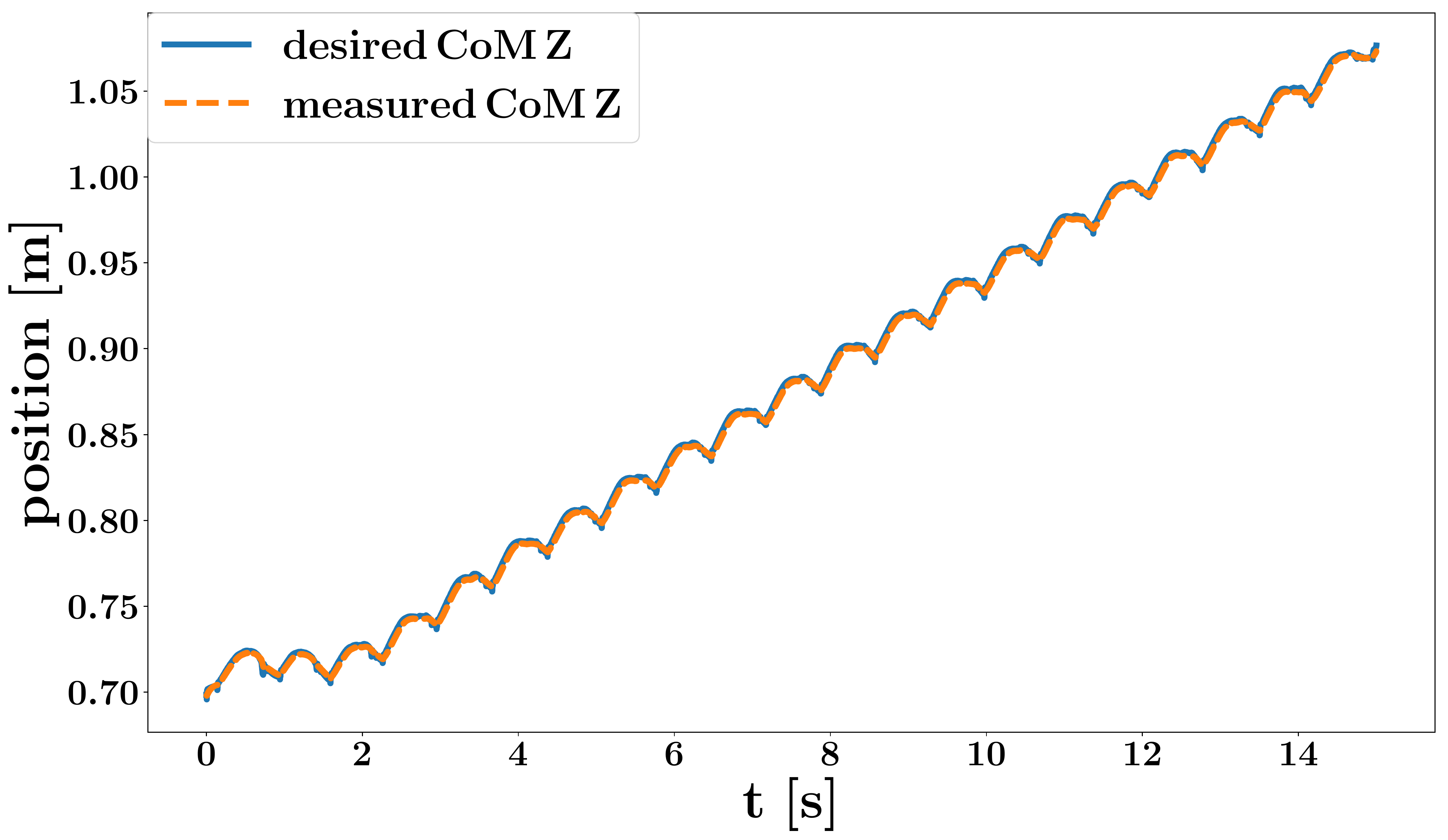}
  \caption{}
  \label{fig:uneven_f}
\end{subfigure} 
\begin{subfigure}{0.22\textwidth}
  \includegraphics[width=1.0\linewidth]{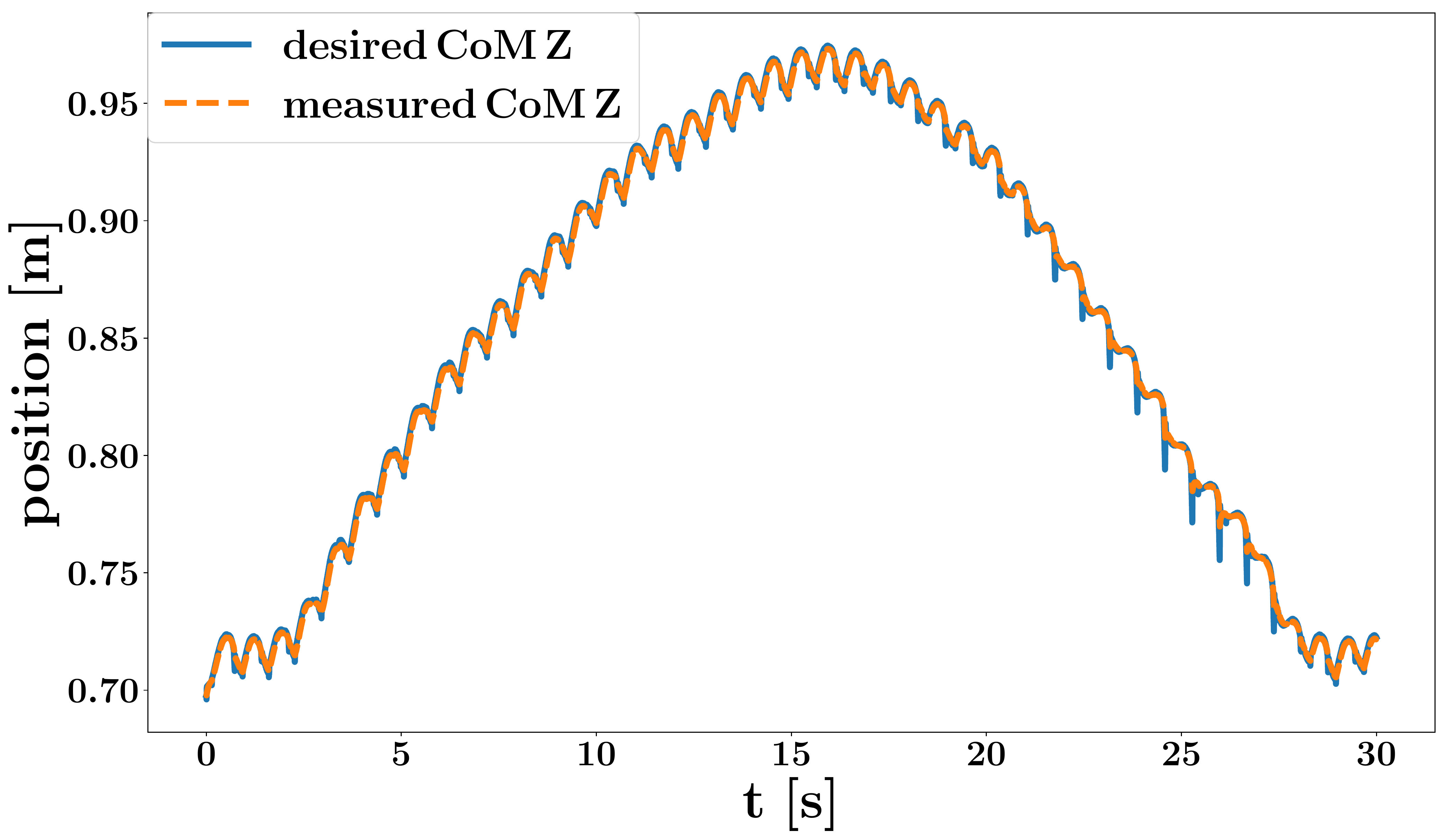}
  \caption{}
  \label{fig:uneven_g}
\end{subfigure}
  \hspace{0.01cm}
\begin{subfigure}{0.22\textwidth}
  \includegraphics[width=1.0\linewidth]{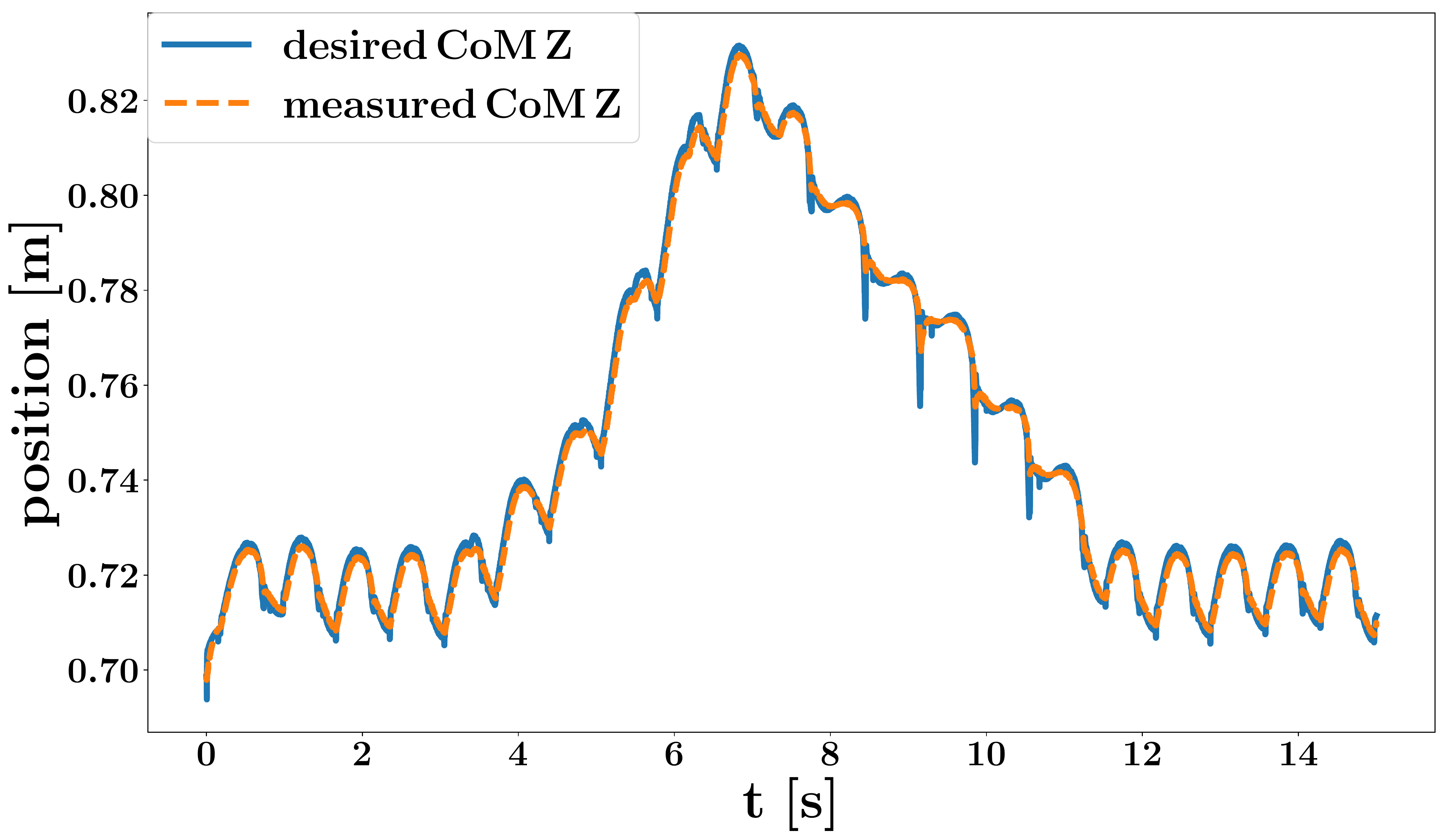}
  \caption{}
  \label{fig:uneven_h}
\end{subfigure} 
\caption{Snapshots and plots of SLIDER blindly walking on different kinds of terrains, with a constant forward velocity of 0.3 m/s ($\text{a}\sim\text{c}$) and 0.6 m/s ($\text{d}$). On top row, from left to right: $\text{(a)}\sim\text{(d)}$: Walk on flat terrain. Walk on a slope with an angle of 15 degrees. Walk on a wave field approximated by slopes with 3 different angles: 15, 10 and 5 degree. Walk over stairs, the successive elevation changes are +2 cm, +2 cm, +3 cm, +3 cm, -2 cm, -3 cm, -2 cm, -3 cm. On bottom row, from left to right, $\text{(e)}\sim\text{(f)}$: The plots of desired and measured CoM $z$ position in corresponding scenarios.}
\label{fig:unevenScenario}
\vspace{-3mm}
\end{figure*}

\subsection{Swing Foot Trajectory Generation}
The swing foot trajectory is generated using a fifth order polynomial. The start and final positions, velocities and accelerations are specified and a parametric quintic curve is generated in $x$, $y$ and $z$ directions. The polynomial in $z$ direction has two halves, with the predefined foot height and zero velocity at the midpoint. The start and final positions in $z$ direction are calculated using the estimated CoM $z$ position at the start of current step subtracted by a constant offset. Because the robot is walking blindly we don't define a trajectory for the foot orientation, and this allows the foot to be compliant to a range of surfaces.
\subsection{Whole Body Controller}
The whole-body controller takes the responsibility of computing the joint torques to achieve the desired motions defined in operational space while respecting a set of constraints. The tasks of interest in this paper are the CoM position, the pelvis orientation, the angular momentum of the robot, the feet positions and orientations. The command of each task is comprised of a desired acceleration as a feedforward term and a state feedback term to stabilize the trajectory.  

The task for the linear motion can be expressed as:
\begin{equation*}
    \bm{J}_{\mathrm{T}}\ddot{\bm{q}} =  \ddot{\bm{x}}^{\mathrm{cmd}} - \dot{\bm{J}}_{\mathrm{T}}\dot{\bm{q}},
\end{equation*}
\begin{equation*}
    \ddot{\bm{x}}^{\mathrm{cmd}} = \ddot{\bm{x}}^{\mathrm{des}} + \bm{K}_{\mathrm{P}}^{\mathrm{pos}}(\bm{x}^{\mathrm{des}} - \bm{x}) + \bm{K}_{\mathrm{D}}^{\mathrm{pos}}(\dot{\bm{x}}^{\mathrm{des}} - \dot{\bm{x}}),
\end{equation*}
where $\bm{J}_{\mathrm{T}}$ is the translational Jacobian for the task and $\bm{x}$ is the actual position of the link. For the task of angular momentum the centroidal momentum matrix \cite{orin2013centroidal} is used as the task jacobian. In uneven terrain walking the angular momentum task is defined  as  a  damping  task  that  damps  out  excess  angular momentum to make the walking robot more stable.

For the task of the angular motion, the command can be formulated as:
\begin{equation*}
    \bm{J}_{\mathrm{R}}\ddot{\bm{q}} =  \dot{\bm{\omega}}^{\mathrm{cmd}} - \dot{\bm{J}}_{\mathrm{R}}\dot{\bm{q}},
\end{equation*}
\begin{equation*}
    \dot{\bm{\omega}}^{\mathrm{cmd}} = \dot{{\bm \omega}}^{\mathrm{des}} + \bm{K}_{\mathrm{P}}^{\mathrm{ang}}(\mathrm{AngleAxis}(\bm{R}^{\mathrm{des}}\bm{R}^T)) +  \bm{K}_{\mathrm{D}}^{\mathrm{ang}}(\bm{\omega}^{\mathrm{des}} - \bm{\omega}),
\end{equation*}
where $\bm{J}_{\mathrm{R}}$ is the rotational Jacobian for the task, $\bm{R}$ and $\bm{R}^{\mathrm{des}}$ denote the actual and desired orientation of the pelvis link respectively, 
$\mathrm{AngleAxis}()$ maps a rotation matrix to the corresponding axis-angle representation, 
$\bm {\omega} \in \mathbb{R}^3$ is the angular velocity of the link. The angular motion is needed because when walking on the uneven terrain, the pelvis orientation needs to be regulated around a nominal orientation to maintain a good posture. We set small values for $\bm{K}_{\mathrm{P}}^{\mathrm{ang}}$ and $\bm{K}_{\mathrm{D}}^{\mathrm{ang}}$ in the foot orientation task to make the ankle compliant so that the foot can adapt to different terrains.

The whole body controller can be formulated as a QP problem below:
\begin{align}
\min_{\ddot{\bm{q}},\, \bm{f}} \quad & \| \bm{A}\ddot{\bm{q}} + \dot{\bm{A}}\dot{\bm{q}} - \bm{X}^{\mathrm{cmd}} \|_{\bm{W}}^2 \tag{1a}\\
\textrm{s.t.} \quad & \bm{M}_f\bm{\ddot{q}} - \bm{J}^{\mathrm{T}}_f\bm{f} = - \bm{H}_f \tag*{(floating base dynamics) \quad (1b) }\\
& \bm{P}\bm{f} \leq \bm{0}\tag*{(pyramidal friction cone) \quad (1c)}\\
& \bm{S}_a^{-1}(\bm{M}_a \ddot{\bm{q}} + \bm{H}_a - \bm{J}^{\mathrm{T}}_a\bm{f}) \in [\bm{\tau}_{min},\, \bm\tau_{max}]\tag*{(input limits) \quad (1d)}
\end{align}
where $\bm{A}$ is a stack of the Jacobian matrices for the tasks of interest, $\bm{X}^{\mathrm{cmd}}$ is a stack of the commanded accelerations and $\bm{W}$ is the weighting matrix, $\bm{P}$ denotes the linearized friction cone. We treat the unilateral contact constraint as a soft constraint by simply assigning a large weight on the desired zero acceleration of the foot \cite{apgar2018fast} \cite{kuindersma2016optimization}. This can speed up the optimization and it is reported in \cite{feng2015optimization} that this gives better stability.

The output joint torque commands $\bm \tau$ at each control iteration can be computed by
\begin{equation}
    \bm\tau = \bm{S}_a^{-1}(\bm{M}_a \bm{\ddot{q}} + \bm{H}_a - \bm{J}^{\mathrm{T}}_a\mathbf{f})
\end{equation}

\begin{figure}[t!]
    \centering
    \begin{subfigure}{0.21\textwidth}
      \includegraphics[width=1.0\linewidth,height=3.0cm]{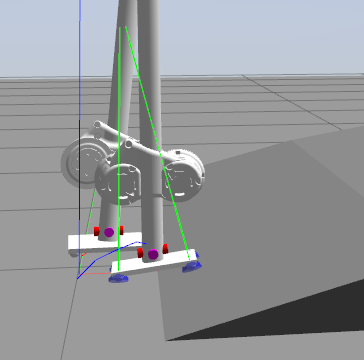}
    \end{subfigure}
      \hspace{0.05cm}
    \begin{subfigure}{0.21\textwidth}
      \includegraphics[width=1.0\linewidth,height=3.0cm]{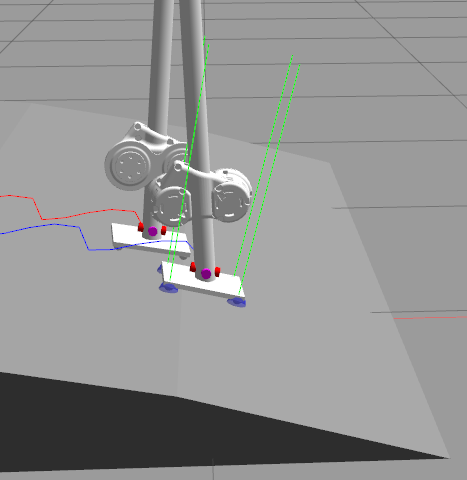}
    \end{subfigure} 
    \caption{With the whole body controller the foot is compliant and can do transitions to different terrains even the robot doesn't know how the terrain looks like. This makes the robot robust to unobserved terrain changes.}
    \label{fig:compliant}
    \vspace{-3mm}
\end{figure}

\section{Results}
This section discusses implementation details and simulation results of SLIDER robot walking on different kinds of uneven terrains including slopes, wave fields and stairs. All the experiments are shown in detail in the accompanying video.
\subsection{Implementation}
Both the high level footstep planner and low level trajectory tracking controller are implemented in C++ for real-time performance. In the whole body controller we use Pinocchio \cite{pinocchioweb} to compute the full rigid body dynamics and qpOASES \cite{ferreau2014qpoases} to solve the related QP problem in both levels. All experiments were carried out in robot simulation environment Gazebo \cite{Koenig04designand} with the physics engine ODE \cite{drumwright2010extending}, using the full dynamics of the real SLIDER robot. The communication through different levels of the control hierarchy is achieved through ROS.

The decoupled aSLIP parameters are set to match with the physical SLIDER robot, here $r_0$ = 0.715~m, m = 14.5~kg, k = 1470~N/m. The step duration is chosen to be 0.7~s and the footheight in the swing foot trajectory generation is 5~cm. The sampling time in the discrete-time MPC is 0.1~s in $x$ and $y$ directions and 0.05~s in $z$ direction. The motion planner predicts 4 steps in horizontal states and 1 step in the vertical state.
We use the same parameters among all walking experiments except that the forward velocity is different. In the stair walking experiment the robot has to walk faster otherwise the foot might hit the edge of the stair.

\subsection{Flat Ground Walking}
We first validate our approach on flat ground. Because there are no changes in $z$ direction on flat ground, the $z$ position of CoM oscillates around $r_0$, as shown in Fig. (\ref{fig:uneven_a})(\ref{fig:uneven_e}).

\subsection{Walking on Smooth Uneven Terrain}
We then validate our approach on slopes and wave fields where the change of terrain height is smooth. With a forward velocity of 0.3~m/s, the SLIDER robot can walk on a slope with 15 degree and a wave field approximated by slopes with 3 different angles: 15, 10 and 5 degrees, as shown in Fig. (\ref{fig:uneven_b}) (\ref{fig:uneven_c})(\ref{fig:uneven_f})(\ref{fig:uneven_g}).
In the experiment the whole body controller plays an important role in making the robot remain robust to unobserved terrain changes. By properly tuning the PD gain of the foot orientation task in the whole body controller, the foot is compliant to adapt to large terrain changes, as shown in Fig. \ref{fig:compliant}.

\begin{figure}[t!]
    \centering
    \begin{subfigure}{0.23\textwidth}
      \includegraphics[width=1.0\linewidth]{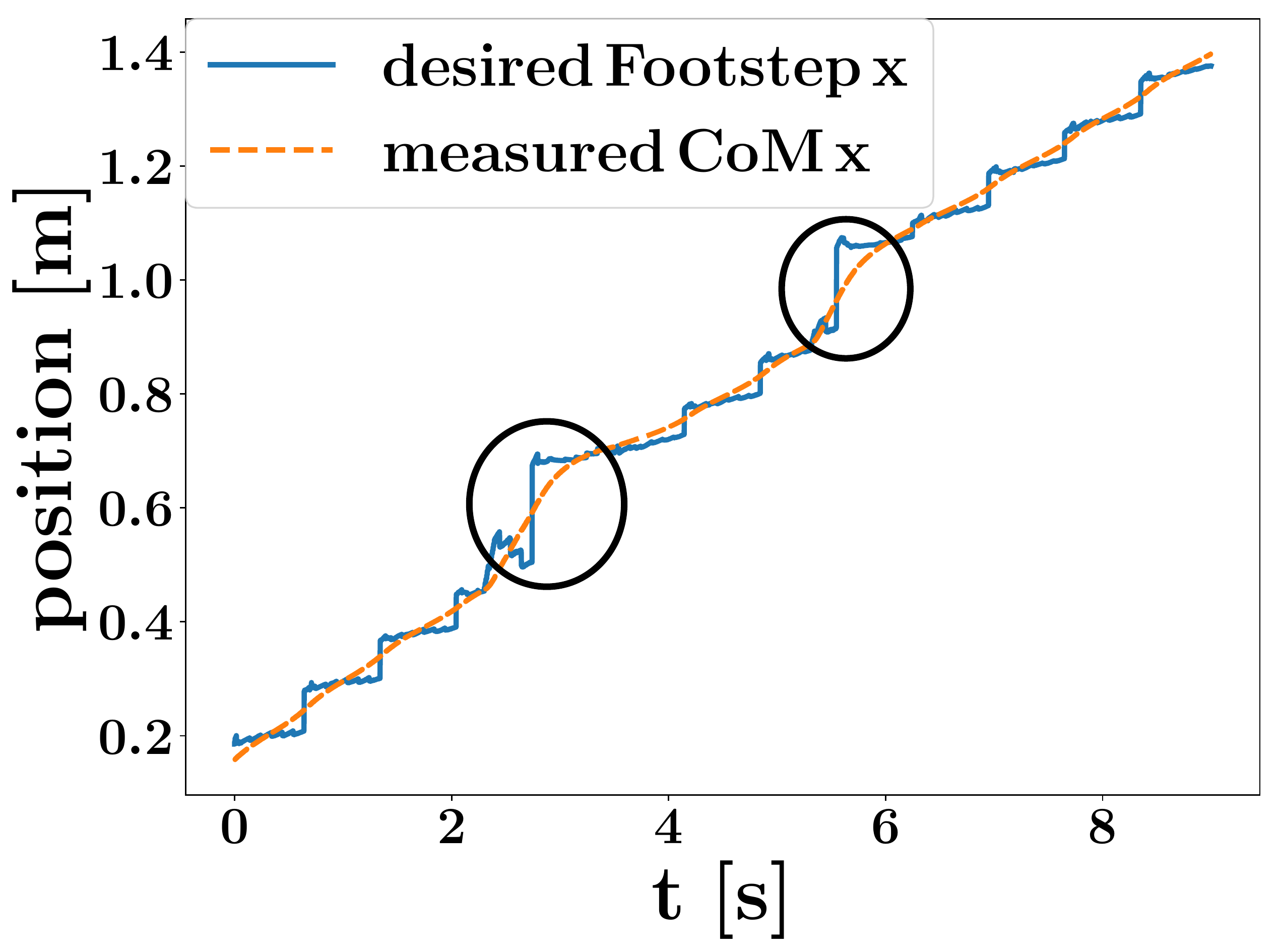}
    \end{subfigure}
      \hspace{0.05cm}
    \begin{subfigure}{0.23\textwidth}
      \includegraphics[width=1.0\linewidth]{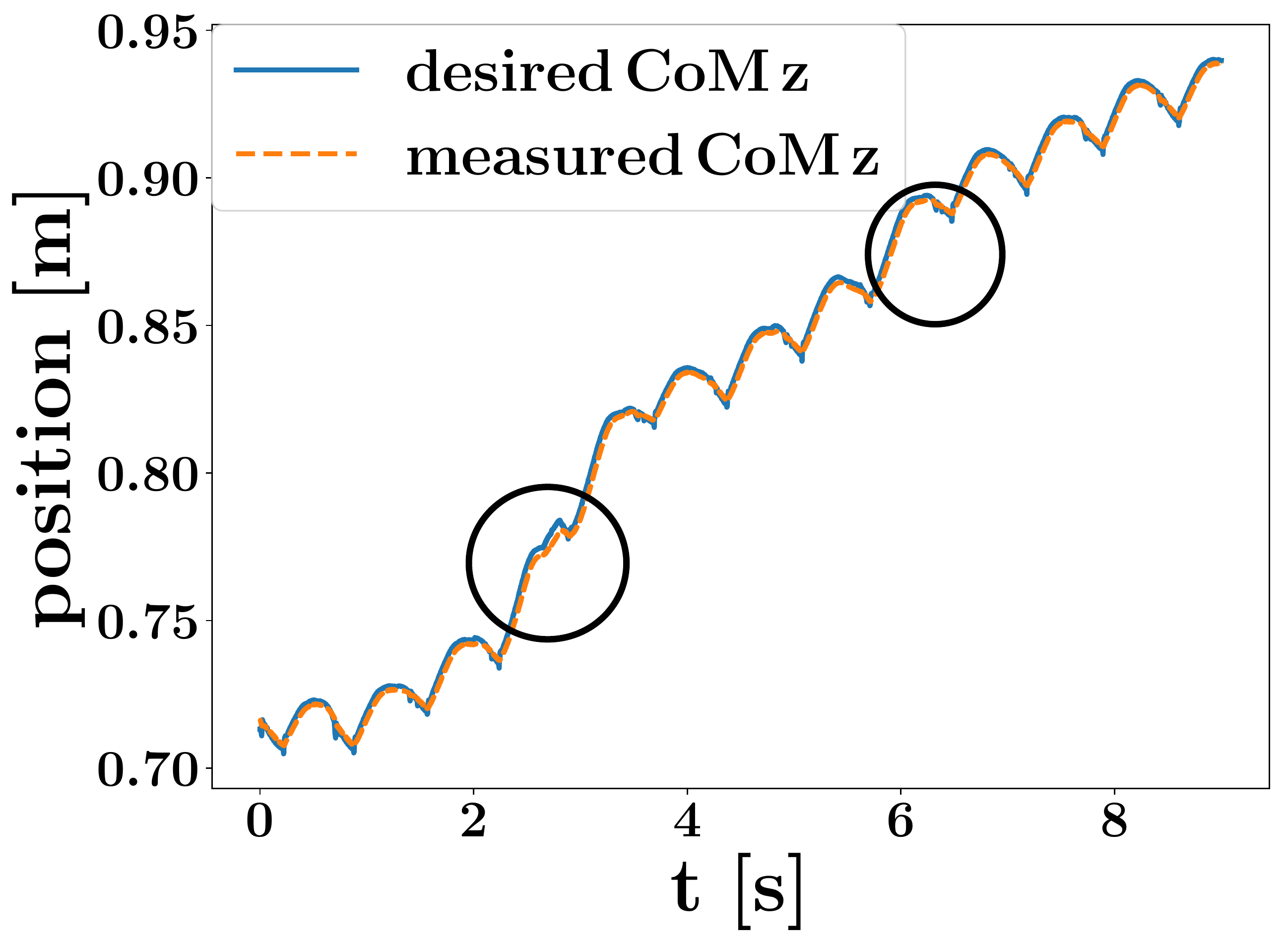}
    \end{subfigure} 
    
    \begin{subfigure}{0.23\textwidth}
      \includegraphics[width=1.0\linewidth]{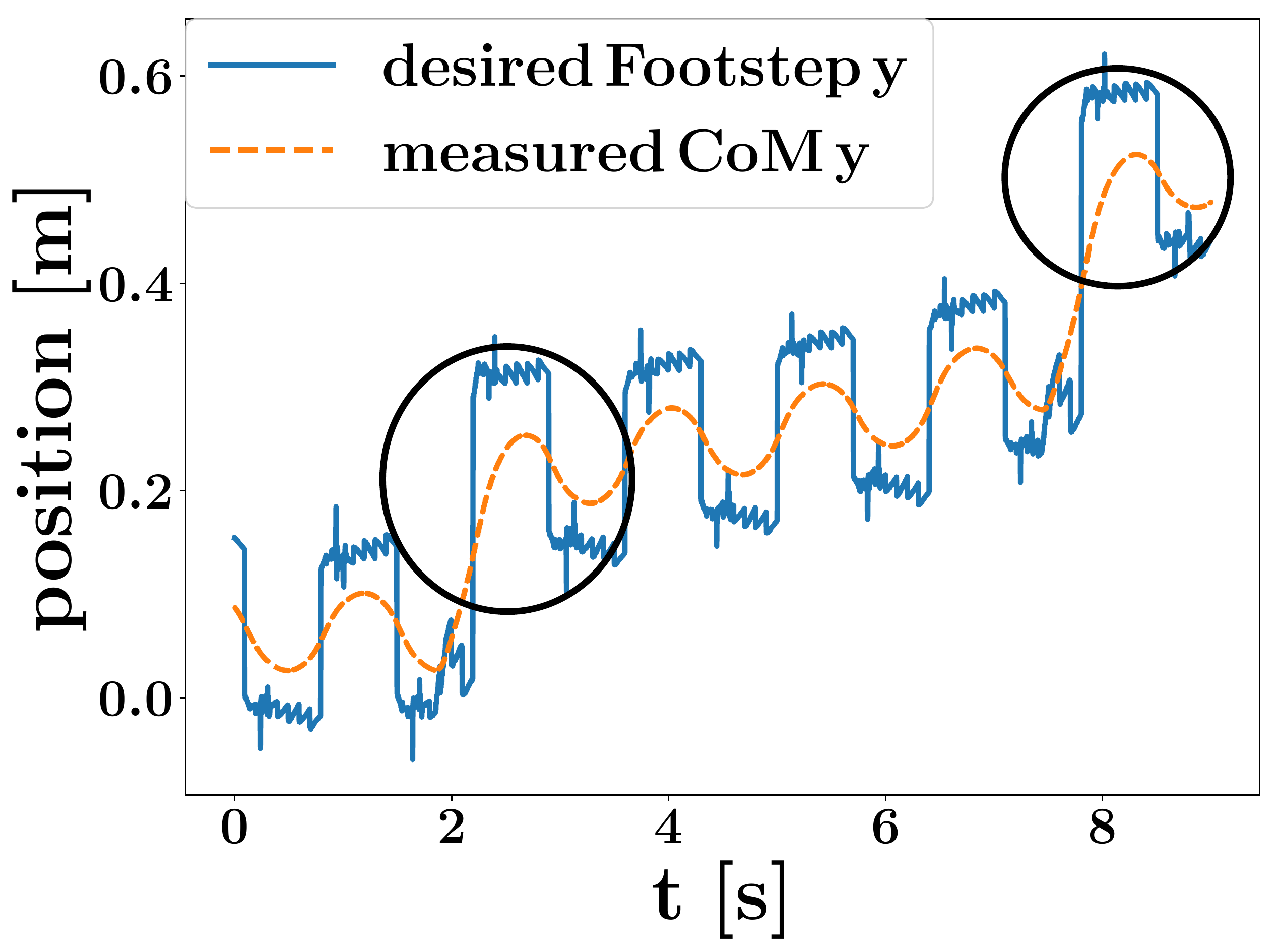}
    \end{subfigure}
      \hspace{0.05cm}
    \begin{subfigure}{0.23\textwidth}
      \includegraphics[width=1.0\linewidth]{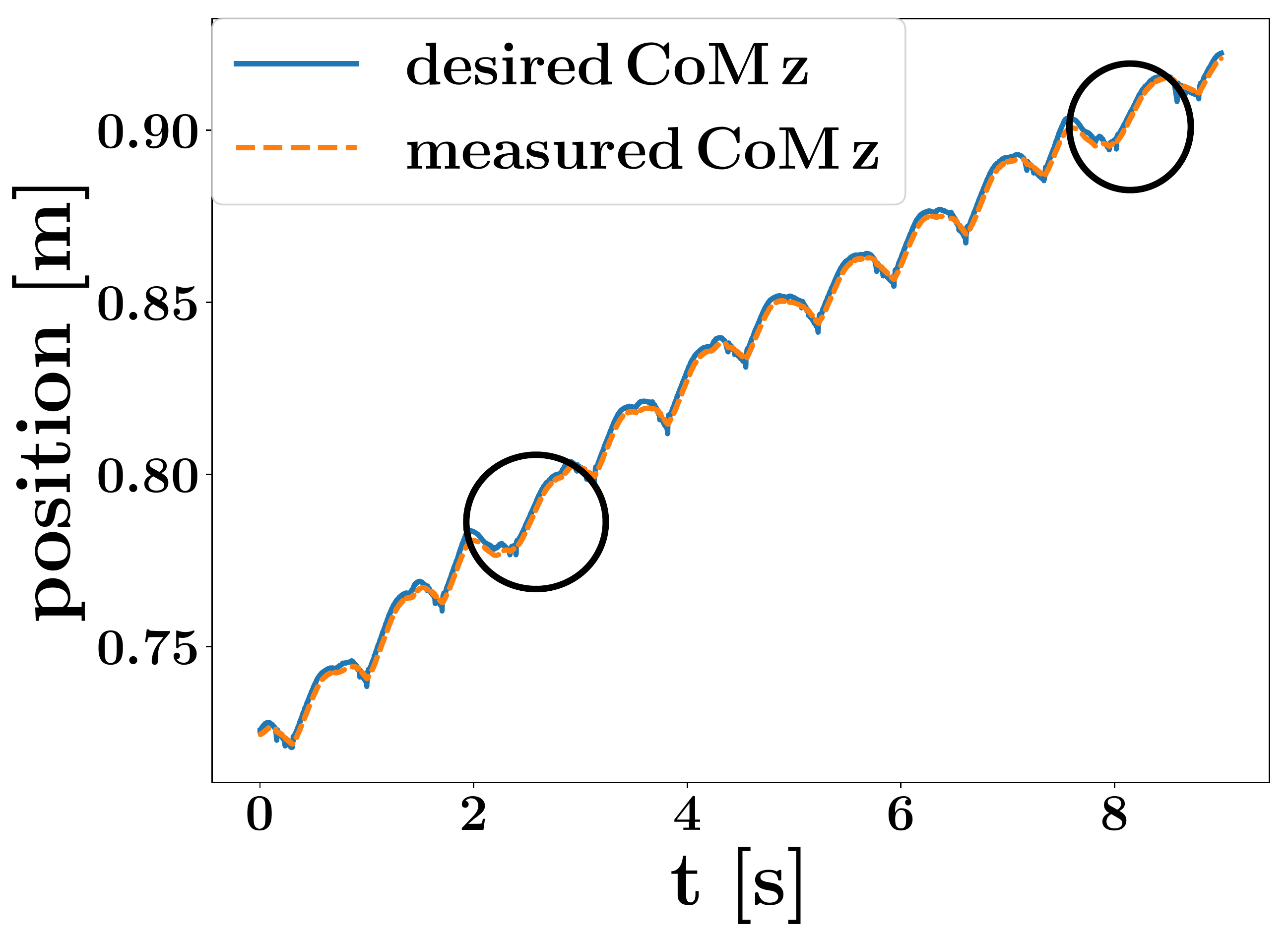}
    \end{subfigure} 
    \caption{Plots showing the states of CoM when the robot got pushed in $x$ or $y$ direction while walking on a wave field at a speed of 0.3~m/s. The pushes are indicated with circles. All impulses were applied with a value of 40~N for a duration of 0.1~s. Top row: states of CoM when got pushed in $x$ direction. Bottom row: states of CoM when got pushed in $y$ direction. }
    \label{fig:push_recovery}
    \vspace{-3mm}
\end{figure}

Further experiments like pushing the robot while it was walking on uneven terrains were performed. As shown in Fig. \ref{fig:push_recovery}, the robot was pushed in $x$ or $y$ direction twice while it walking on a wave field. When the robot was pushed in $x$ direction, the robot quickly took a large step in $x$ direction to regulate the CoM velocity back to the desired velocity. There is also a big change in $z$ direction because the wave field is ascending in $x$ direction when the robot got pushed, but the motion quickly got stabilized. In the case of $y$ direction push, the robot also took a large step in $y$ direction and stabilized in one step. As the terrain height is only changing along $x$ direction, there is no big change in $z$ direction in this case.

\subsection{Walking on Discrete Uneven Terrain}
Discrete uneven terrain provides bigger instantaneous changes to terrain height compared with smooth uneven terrain. As shown in Fig. (\ref{fig:uneven_d}), the robot walked blindly over a set of stairs with a biggest height change of 3~cm. The robot walked with a forward velocity of 0.6~m/s and a foot height of 5~cm in swing foot phase. As shown in Fig. (\ref{fig:uneven_h}) the CoM Z position exhibits large variations when walking over stairs but was stabilized quickly, which demonstrates the robustness of our approach. In our experiment the highest steps the robot can walk over is 3~cm, for higher steps the foot would hit the edge of the stair and get stuck. 

\section{DISCUSSION}
The proposed methods is successfully applied to SLIDER robot and demonstrates its performance of 3D blind walking on various uneven terrain and robustness to disturbances. Because of SLIDER's lightweight legs, the robot can perform fast leg movements which helps to stable the robot. However the proposed method is general and can be applied to various legged robots.

There are other techniques dealing with variable height CoM
trajectories \cite{englsberger2013three, 8206385}. But these techniques either requires the controller to be terrain-aware to plan future CoM motion or having nonlinear dynamics so a solution is not guaranteed. Our proposed controller keeps the average CoM height constant in one step by following a spring dynamics. This has two advantages: firstly the controller doesn't require terrain information, secondly the spring dynamics enables the vertical compliance of the robot so that the robot is robust to unexpected height variations.  

The author observed that the tracking error is larger when the robot walks down the wavefield or stairs than walking up, as shown in Fig.\ref{fig:uneven_g}, Fig. \ref{fig:uneven_h}. This happens because the feet are still in the air at the end of one step when the robot walks down. The unexpected sudden drop of the feet at the start of next step gives the robot a large impact. For robot walking upwards, the effect of early touchdown can be alleviated by the compliant feet and vertical compliance of the CoM. To improve, a controller with contact detection or terrain information can achieve a smaller tracking error.  
\section{CONCLUSIONS AND FUTURE WORK}

We present a highly reactive controller which enables robots to blindly walk over various kinds of uneven terrains while resisting pushes. The high level motion planner performs fast online optimization on footstep locations and CoM height, and the low level inverse-dynamics based whole body controller tracks the trajectory. We show in simulation that using this controller, the robot, SLIDER, can walk over slopes, wave fields and stairs without any terrain information and can also recover from pushes while walking. Future work will involve implementing our approach on the real SLIDER robot, and also incorporating perception information into the high level step planner.

\addtolength{\textheight}{-9cm}   





\section*{ACKNOWLEDGMENT}
 Ke Wang is funded by the CSC Imperial Scholarship. The authors would thank Digby Chappell for meaningful discussions.

\bibliographystyle{IEEEtran}
\bibliography{bibliography.bib}
\end{document}